\documentclass[letterpaper, 10pt, conference]{ieeeconf}
\IEEEoverridecommandlockouts 
\usepackage{amsmath,amssymb,url}
\usepackage{graphicx,subfigure}
\usepackage{color}
\usepackage{siunitx}
\usepackage[hidelinks]{hyperref}
\usepackage{cleveref}
\usepackage{booktabs}

\usepackage{tikz}
\usepackage{mathdots}
\usepackage{yhmath}
\usepackage{cancel}
\usepackage{color}
\usepackage{siunitx}
\usepackage{array}
\usepackage{multirow}
\usepackage{amssymb}
\usepackage{gensymb}
\usepackage{tabularx}
\usepackage{extarrows}
\usepackage{booktabs}
\usetikzlibrary{fadings}
\usetikzlibrary{patterns}
\usetikzlibrary{shadows.blur}
\usetikzlibrary{shapes}

\newcolumntype{?}{!{\vrule width 1pt}}

\newcommand{\SO}{\ensuremath{\mathsf{SO(3)}}}

\renewcommand{\Re}{\ensuremath{\mathbb{R}}}

\title{\LARGE \bf
    Vision-in-the-loop Simulation for Deep Monocular Pose Estimation of UAV in Ocean Environment
}
\author{Maneesha Wickramasuriya, Beomyeol Yu, Taeyoung Lee, and Murray Snyder
	\thanks{Maneesha Wickramasuriya, Beomyeol Yu, Taeyoung Lee, and Murray Snyder, Mechanical and Aerospace Engineering, George Washington University, Washington, DC 20051, {\tt \{maneesh,yubeomyeol,tylee,snydermr\}@gwu.edu}}%
    \thanks{\textsuperscript{\footnotesize\ensuremath{*}}This research has been supported in part by USNA/NAVSUP (N0016123RC01EA5), AFOSR MURI (FA9550-23-1-0400), and ONR (N00014-23-1-2850).}
}

\graphicspath{{./figs/}}

\makeatletter
\def\endthebibliography{%
	\def\@noitemerr{\@latex@warning{Empty `thebibliography' environment}}%
	\endlist
} 
\makeatother

\begin{document}
\allowdisplaybreaks
\maketitle \thispagestyle{empty} \pagestyle{empty}

\begin{abstract}
    This paper proposes a vision-in-the-loop simulation environment for deep monocular pose estimation of a UAV operating in an ocean environment. 
    Recently, a deep neural network with a transformer architecture has been successfully trained to estimate the pose of a UAV relative to the flight deck of a research vessel, overcoming several limitations of GPS-based approaches. 
    However, validating the deep pose estimation scheme in an actual ocean environment poses significant challenges due to the limited availability of research vessels and the associated operational costs.
    To address these issues, we present a photo-realistic 3D virtual environment leveraging recent advancements in Gaussian splatting, a novel technique that represents 3D scenes by modeling image pixels as Gaussian distributions in 3D space, creating a lightweight and high-quality visual model from multiple viewpoints. 
    This approach enables the creation of a virtual environment integrating multiple real-world images collected in situ.
    The resulting simulation enables the indoor testing of flight maneuvers while verifying all aspects of flight software, hardware, and the deep monocular pose estimation scheme. 
    This approach provides a cost-effective solution for testing and validating the autonomous flight of shipboard UAVs, specifically focusing on vision-based control and estimation algorithms.
\end{abstract}

\section{Introduction}
Unmanned Aerial Vehicles (UAVs) have become integral to modern technology, with applications in military operations, maritime surveillance, search and rescue, and offshore logistics.
Despite their advantages—such as cost-effectiveness, versatility, and enhanced safety—operating UAVs in complex and dynamic environments like oceans presents significant challenges.
Autonomous takeoff, navigation, and landing on moving platforms such as ships remain particularly difficult due to unpredictable maritime conditions, including ship motion, variable lighting, and air wakes \cite{JGCD_Ship_Airwakes}.

A crucial factor in enabling safe UAV operations on ships is accurately estimating the UAV’s relative 6D pose (position and orientation in 3D space) with respect to the vessel.
Traditional methods often rely on Real-Time Kinematic Global Positioning System (RTK-GPS), which provides precise positioning through differential corrections.
However, GPS-based systems have limitations, including dependence on continuous communication between the UAV and the ship, susceptibility to signal jamming or spoofing, and the risk of revealing the ship’s location through radio transmissions \cite{JGCD_RTKGPS_Visual_marker}.
These constraints highlight the need for alternative approaches, such as vision-based methods that use monocular cameras for relative pose estimation.

Vision-based methods—including fiducial markers like ArUco \cite{Aruco2}, laser patterns, and feature-based visual-inertial navigation (VIN)~\cite{KaniDelKal}—have been explored for UAV landing operations.
While effective under controlled conditions, these methods are sensitive to occlusions, lighting variations, and visibility constraints, limiting their robustness in maritime environments.
To address these challenges, prior research demonstrated that deep learning-based approaches using a Transformer Neural Network Multi-Object (TNN-MO) architecture outperform conventional methods in estimating a UAV’s relative pose using monocular images \cite{wick2024}.
This approach improved accuracy in both synthetic and real-world datasets, overcoming several limitations of traditional techniques.

The UAV-mounted camera captures images of the real ocean environment, which are then processed by the TNN-MO model to estimate the camera’s relative 6D pose with respect to the ship.
However, testing and validating these deep learning models in actual ocean environments poses significant challenges due to the limited availability of research vessels, high operational costs, and the risk of UAV failures—potentially leading to costly equipment losses and further complicating research efforts.

To address these challenges, this paper presents a vision-in-the-loop simulation environment for testing deep monocular pose estimation models and vision-based control algorithms.
The framework leverages advancements in Gaussian splatting, a novel technique for photorealistic 3D scene reconstruction that represents a scene using 3D Gaussian functions instead of traditional meshes or point clouds~\cite{kerbl3Dgaussians},
which enables real-time rendering of highly detailed and realistic environments.
By constructing a photorealistic 3D virtual environment, this approach enables cost-effective evaluation of flight software and hardware under realistic maritime conditions in indoor lab settings, reducing reliance on research vessel deployments or scaled-down indoor vessel prototypes.

In short, the main contribution of this paper is the development of a UAV test framework that enables comprehensive indoor validation of UAV deep pose estimation methods and autonomous flight maneuvers, addressing key challenges in real-world testing.
By integrating advanced simulation techniques with deep learning, this study enhances the robustness and reliability of UAV operations in ocean environments.

While this study focuses on maritime applications, the framework’s key advantage—its ability to generate photorealistic environments with ease and cost efficiency—makes it highly adaptable to various real-world visual conditions.
This flexibility allows researchers to replicate diverse operational scenarios, each with unique challenges, ensuring broader applicability across multiple domains.

The remainder of this paper is organized as follows.
\Cref{sec:TNN} discusses the deep transformer network architecture for monocular pose estimation and describes the generation of synthetic data for model training and validation.
\Cref{sec:PRVE} introduces the proposed photorealistic virtual environment based on Gaussian splatting, detailing the data collection and visualization processes.
\Cref{sec:Sim} outlines the vision-in-the-loop simulation framework, including the integration of flight hardware and software.

\section{Deep Transformer Network for Monocular Pose Estimation}\label{sec:TNN}
To estimate the 6D pose of UAV from a monocular camera, we employed the Transformer Neural Network for Multi-Object Pose Estimation (TNN-MO) architecture proposed in our previous work~\cite{wick2024}.
This architecture is specifically designed to estimate 6D poses (position and orientation) from a single RGB image with high accuracy and robustness in complex, real-world environments.
\subsection{Network Architecture}

The TNN-MO architecture combines a CNN, a transformer module, and probabilistic fusion for precise pose estimation.
It processes a $480\times640$ single red, green, blue (RGB) image through a ResNet50 backbone, reducing the resolution to $15\times20$ and generating a 2048-channel feature map.
A $1\times1$ convolution reduces the number of channels to 256.
These features are flattened, enriched with positional encodings, and processed by a 6-layer transformer encoder with skip connections.
A 6-layer transformer decoder, using 7 object queries (for 6 ship parts and a no-object class), outputs embeddings for class prediction and 2D keypoints via FFNs.
By estimating 2D keypoints, the model recovers the relative 6D poses of objects using the Efficient Perspective-n-Point (EPnP)~\cite{EPnP} algorithm.
EPnP solves the 2D-to-3D correspondence problem by leveraging known 2D keypoints and their associated 3D object models, enabling accurate estimation of each object’s relative position and orientation.
Finally, Bayesian fusion integrates pose estimates from multiple ship parts, weighting contributions by confidence to enhance robustness and accuracy.

\subsection{Synthetic Data and Model Training}

To ensure effective neural network training, we developed a synthetic data generation pipeline within a simulated virtual environment, eliminating the cost and time associated with real-world data collection.
A detailed 3D CAD model of a research vessel was constructed using structure-from-motion techniques, with missing elements manually added to enhance fidelity.
Environmental features such as dynamic ocean waves, sunlight, shadows, and human figures were incorporated to improve realism.
Domain randomization was applied to textures (e.g., sea, sky, hull) and lighting conditions to ensure the model generalizes across diverse scenarios.
Camera poses were sampled from realistic UAV trajectories, capturing various viewpoints and distances.
The dataset included both horizon and non-horizon scenarios to improve robustness under challenging conditions.
To mitigate occlusion issues and enhance detection accuracy, the ship was decomposed into six parts (e.g., stern, superstructure).
The model was trained on 435k synthetic images (332k with horizon, 102k without) using the TNN-MO architecture in PyTorch.
Training spanned 350 epochs with the AdamW optimizer, a batch size of 48, and gradient clipping at a maximum norm of 0.1, leveraging an NVIDIA A100 GPU.

\subsection{Validation}
The TNN-MO model's performance was validated using synthetic and real-world datasets to ensure robustness in estimating a UAV's 6D pose relative to a ship.

\subsubsection{Validation with Synthetic Data}
\begin{figure}[!b] \centering \includegraphics[width=1\linewidth]{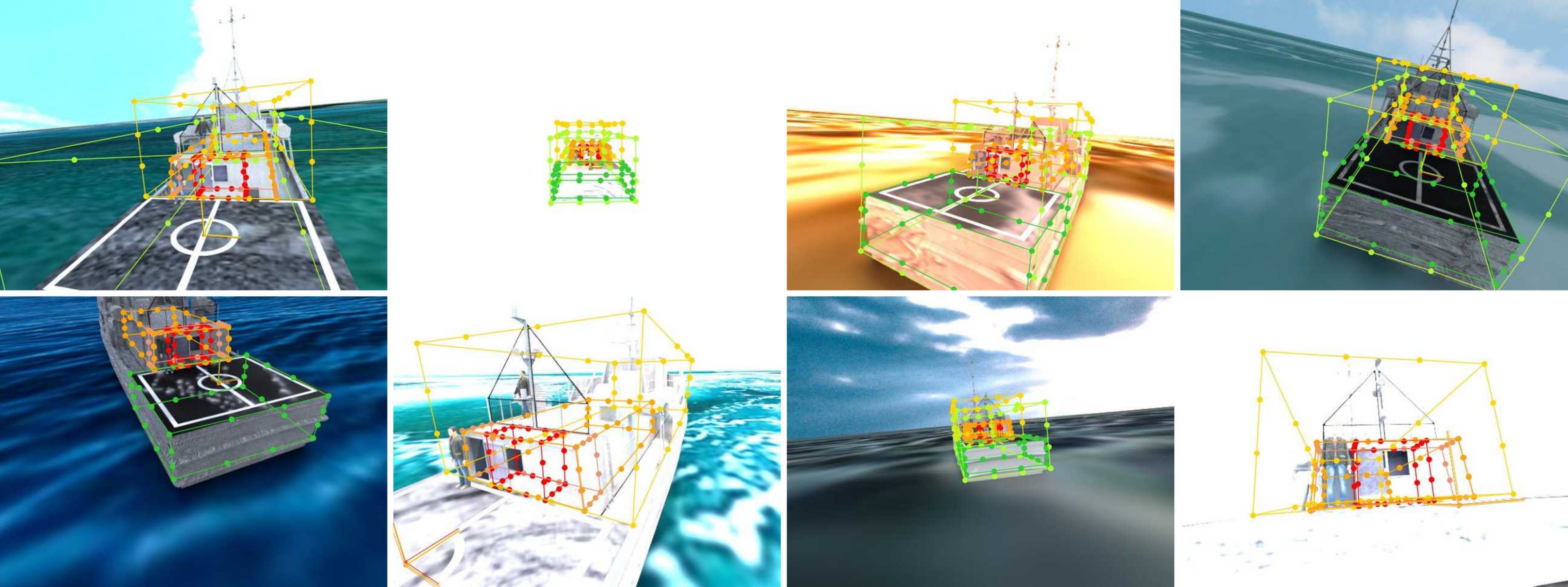} 
	\caption{TNN-MO model tested on synthetic data: for object classes with confidence scores greater than 0.9, the keypoints and base frame are illustrated in different colors.} 
	\label{fig:3} 
\end{figure}

The model was tested on a synthetic dataset, as shown in \Cref{fig:3}, which was excluded from training.
It achieved a mean absolute position error of \qty{0.204}{\meter} and an attitude error of \qty{0.91}{\degree} across 5,500 images under varying conditions, and the position error is about 0.8\% of the maximum range (\qty{25}{\meter}).

\begin{figure}[!hb]
	\centering
	\includegraphics[width=1 \linewidth]{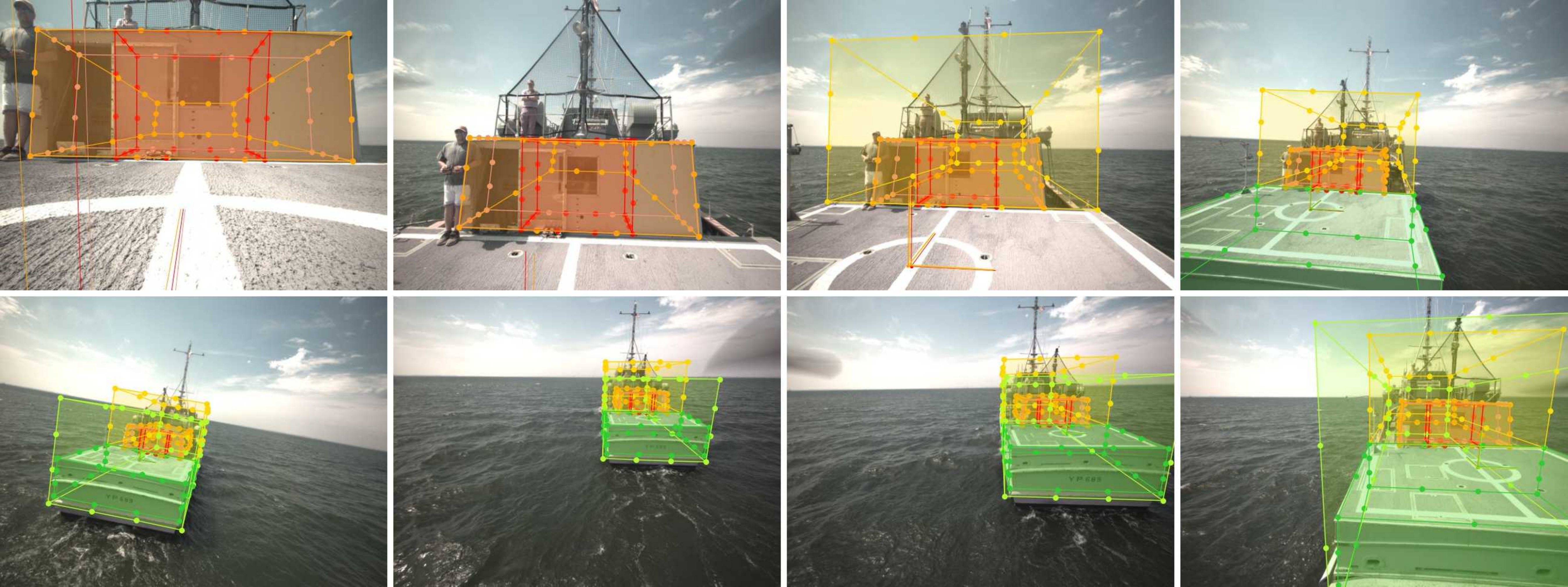}
	\caption{TNN-MO model tested on real-world data: for object classes with confidence scores greater than 0.9, the keypoints and base frame are highlighted in different colors.}
	\label{fig:4}
\end{figure}
\subsubsection{Real-World Validation}
\begin{figure}[t]
	\centering
	\includegraphics[width=1\linewidth]{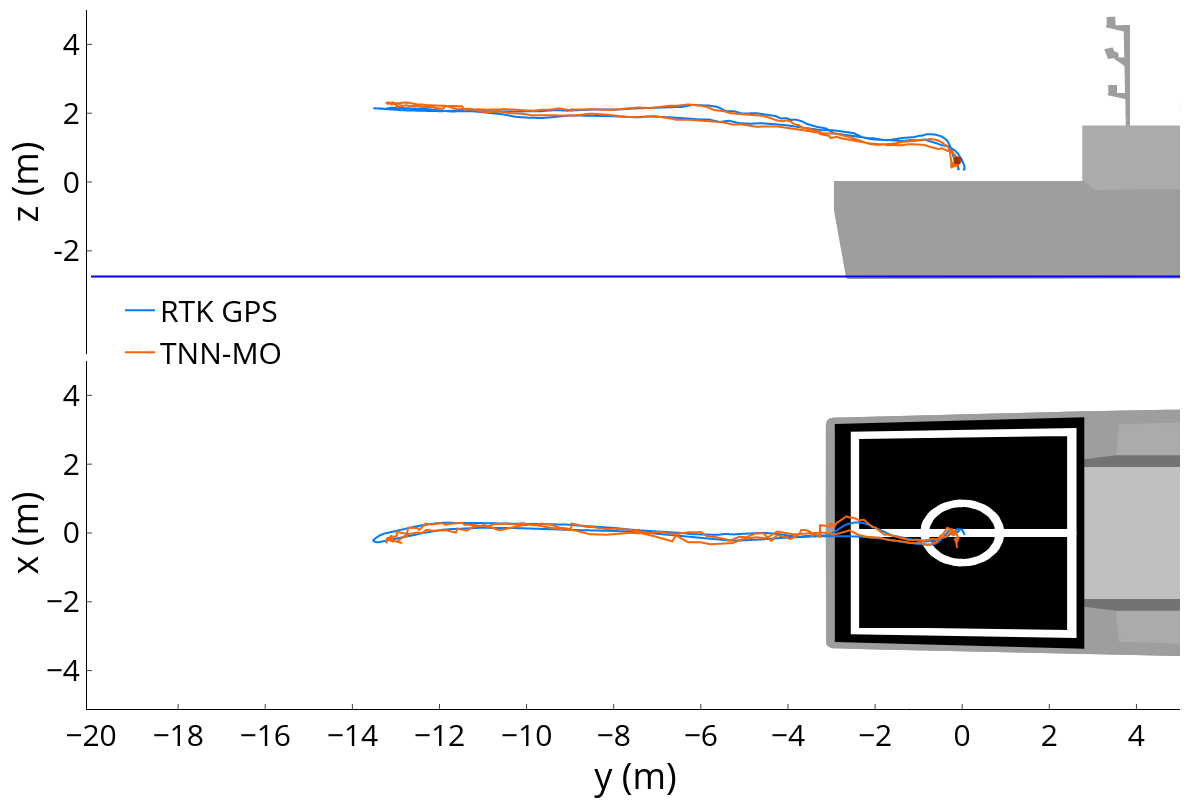}
	\caption{Position estimation for overexposed real-world images: the estimated position (orange) is compared against the RTK-GPS measurements (blue).}
	\label{fig:5}
\end{figure}

Real-world validation, as shown in \Cref{fig:4}, was conducted using a UAV equipped with a Data Collection System (DCS) on the research vessel YP689. 
The model handled variable lighting, occlusions, and partial visibility, achieving position MAE between \qty{0.089}{\meter} and \qty{0.177}{\meter} (0.66\%–0.97\% of \qty{18.2}{\meter} range) and rotation errors of \qty{1.1}{\degree} to \qty{4.0}{\degree}. 
Results aligned with RTK-GPS measurements, as shown in \Cref{fig:5}, demonstrating robustness for autonomous UAV operations.

\section{Photo-Realistic Virtual Environments}\label{sec:PRVE}

The presented TNN-MO model is capable of estimating the 6D pose of a UAV in an ocean environment, as validated by real-world data. This suggests its potential to enable vision-based autonomous UAV flights, ensuring safe and reliable launch and recovery from a research vessel.
However, developing a fully autonomous UAV system that utilizes deep monocular pose estimation models in real-world ocean environments requires substantial additional engineering effort. 
This includes integration with estimation and control architectures, flight hardware and software development, tuning, testing, and validation.
In practice, these tasks pose additional challenges due to the limited availability of research vessels, high operational costs, and the risk of UAV failures, which could lead to costly equipment losses and further complicate research efforts.

Rather than developing these algorithms directly in real-world ocean environments, a more practical approach is to construct an indoor setup that emulates the considered scenarios. 
Traditionally, indoor setups involve building a downscaled ship prototype and simulating environmental conditions to resemble actual ocean scenarios.
However, achieving realistic visual conditions for camera systems and scaling the model accurately are particularly difficult due to lighting constraints, a challenge further exacerbated when the available flight space is significantly limited.

Another widely used approach is to generate photorealistic computer-simulated environments using 3D graphics software.
However, constructing an accurate 3D model of a ship is challenging due to the confidential nature of ship designs, and creating a realistic ocean environment requires specialized expertise in computer graphics.
Additionally, emulating various outdoor conditions at sea is a key challenge.
While the ship model itself may remain unchanged, ocean conditions and lighting vary significantly depending upon the weather, time of day, and season.
Incorporating these dynamic changes into a simulated environment is both complex and computationally demanding.
Real-time rendering of such scenes requires significant graphical processing power and optimization to generate images efficiently.

To address these challenges, we propose a novel simulation framework that leverages recent advancements in Gaussian splatting to generate photorealistic images in real time for real-world pose estimation.
This approach aims to overcome the limitations of traditional methods and provide a more efficient and realistic testing environment for autonomous UAV flight algorithms.

\subsection{Gaussian Splatting} \label{sec:GS}

Gaussian splatting is a computer graphics technique for real-time rendering of 3D radiance fields, offering a bridge between traditional point-based rendering and neural radiance field methods~\cite{kerbl3Dgaussians}. 
It uses 3D Gaussians as primitives for scene representation, optimizing properties such as position, covariance, opacity, and spherical harmonics (SH) coefficients to enhance rendering quality.
These 3D Gaussians provide a differentiable volumetric representation that maintains computational efficiency by avoiding redundant calculations in empty spaces, enabling a compact, high-quality representation without relying on neural networks.

A key advantage of Gaussian splatting is its efficient, visibility-aware rendering algorithm, which leverages a tile-based rasterizer. 
This approach supports anisotropic Gaussian splatting, allowing for fast scene optimization and rendering.
By incorporating adaptive density control, it dynamically adjusts the number and distribution of Gaussians, enabling precise representation of complex geometries. 
The method achieves state-of-the-art visual quality while maintaining real-time performance, rendering at over 30 frames per second (FPS) at 1080p resolution.
These capabilities make Gaussian splatting highly suitable for novel-view synthesis in intricate environments.

Given its strengths, Gaussian splatting is particularly advantageous for generating photorealistic indoor environments to simulate real-world conditions for autonomous UAV testing. 
Unlike traditional methods or computationally intensive 3D graphics pipelines, Gaussian splatting enables the creation of dynamic and realistic environments—such as ocean simulations—with minimal computational overhead.
This real-time rendering capability ensures a high-fidelity testing environment for UAV algorithms, including those integrating deep monocular pose estimation models. 
Consequently, Gaussian splatting provides an efficient, flexible, and repeatable solution for indoor experimentation, overcoming the limitations of traditional setups and enabling precise validation of autonomous UAV flight systems.

\subsection{Data Collection} \label{sec:Data}
\begin{figure*}[t]
	\centering
	\scalebox{0.75}{
		\begin{tikzpicture}[x=0.75pt,y=0.75pt,yscale=-1,xscale=1]
			\draw (350.42,129.83) node  {\includegraphics[width=519.75pt,height=259.37pt]{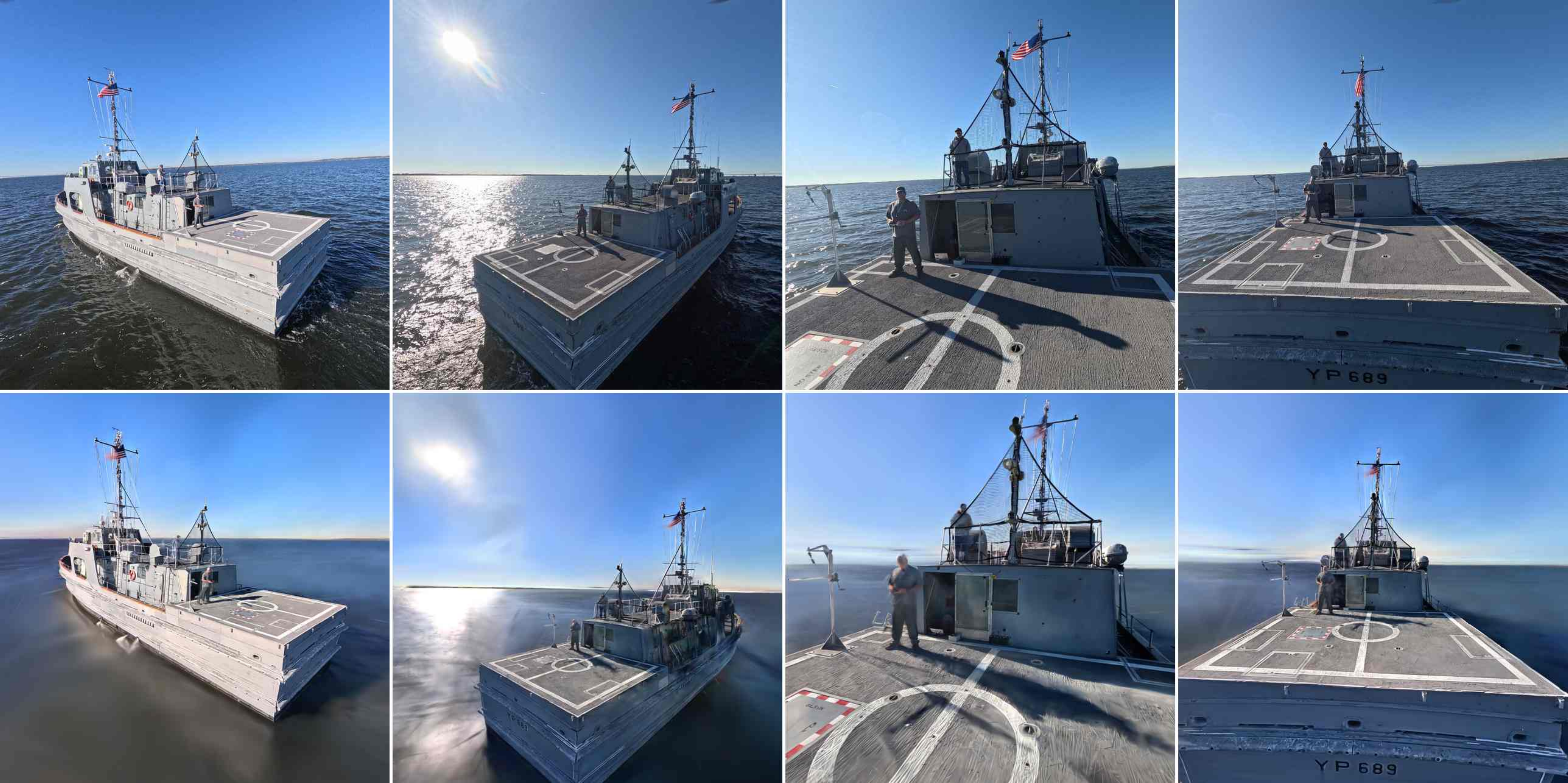}};
			
			\draw (5.92,-40.08) node [anchor=north west][inner sep=0.75pt]   [align=left] {\textbf{{\large \textcolor[rgb]{0.99,0.99,0.99}{Real}}}};
			
			\draw (5,132) node [anchor=north west][inner sep=0.75pt]  [color={rgb, 255:red, 101; green, 31; blue, 31 }  ,opacity=1 ] [align=left] {\textbf{{\large \textcolor[rgb]{1,1,1}{Synthetic}}}};
			
		\end{tikzpicture}
	}
	\caption{Comparison of real-world images captured by the GoPro Hero 13 Black camera (first row) and photo-realistic images generated from the 3D Gaussian Splatting (3DGS) model (second row).}
	\label{fig:real_vs_3DGS}
\end{figure*}
\begin{figure}[t]
	\centering
	\includegraphics[width=0.8 \linewidth]{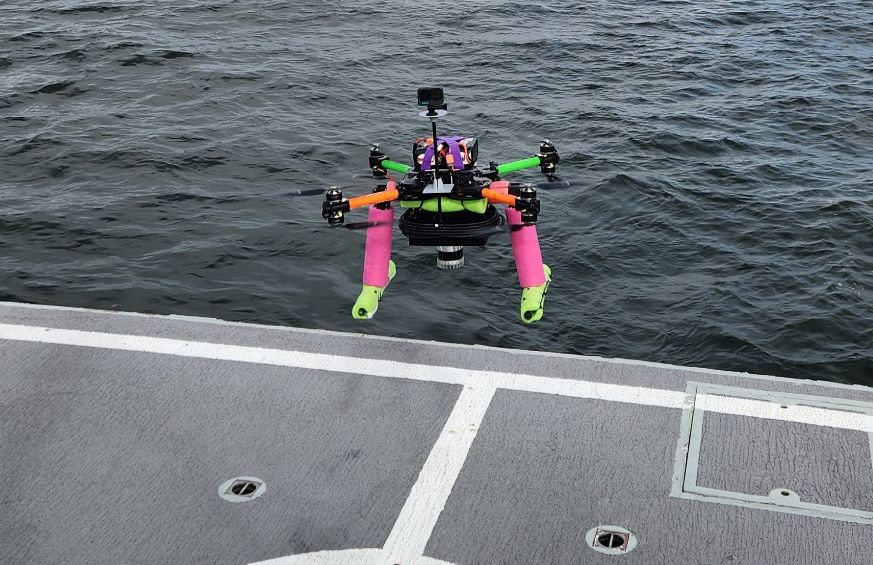}
	\caption{GoPro camera mounted on an octocopter UAV manually flown around a USNA research vessel to collect optical data in Chesapeake Bay, Maryland.}
	\label{fig:6}
\end{figure}


To collect real-world image data for constructing 3D Gaussian Splatting (3DGS), we utilized a GoPro Hero 13 Black action camera mounted on the octocopter UAV, as shown in \Cref{fig:6}. This UAV was previously used for deploying the DCS~\cite{wick2024conf}.
Initially, 4K video footage was recorded with image stabilization enabled. However, generating images from the video introduced motion blur, which significantly degraded the quality of the 3DGS reconstruction.

To mitigate motion blur and improve image quality, we switched to the time-lapse mode, enabling the camera to capture high-resolution images (5546 × 4851 pixels) at fixed intervals of 0.5 seconds.
This approach ensured sharper still frames, thereby enhancing the accuracy and reliability of the 3DGS reconstruction process.

The UAV was flown closer to the ship to capture detailed data, focusing on the back and both sides of the ship, as illustrated in the first row of \Cref{fig:real_vs_3DGS}.
Particular attention was given to the ship’s landing area to ensure comprehensive coverage of the flight deck, thereby improving the robustness of the 3DGS reconstruction, especially near the stern of the ship where the UAV operates more often. 

Image datasets were captured across multiple days, at varying times of day, and under different vessel headings to construct diverse 3D Gaussian Splatting (3DGS) environments with distinct lighting conditions. 
To construct 3DGS environment 1 shown in \Cref{fig:real_vs_3DGS}, a dataset of 1,084 images was used.
During the process, the images were downscaled to a resolution of 1600 × 1399 pixels to manage GPU memory limitations.
The reconstruction was performed using an NVIDIA A100-PCIE-80GB GPU, ensuring sufficient computational power for efficient processing.

\subsection{Virtual Ocean Environment} \label{sec:Viz}

After training the 3D Gaussian Splatting (3DGS) model using the collected GoPro data, we successfully generated photo-realistic virtual environments with different light conditions.
A comparison between the real-world photos and the virtual images generated by the proposed 3DGS is presented in \Cref{fig:real_vs_3DGS}, which demonstrates that the virtual scene closely resembles the actual environment, achieving a high level of realism.
Specifically, the sky and horizon are rendered realistically.
However, the ocean appears blurry in the generated images due to the inherent noise in feature extraction caused by non-stationary ocean waves.
Despite this, we trained the model to prioritize the ship’s structure while minimizing the impact of the ocean and sky by employing domain randomization techniques \cite{wick2024}.
As a result, the ocean would have minimal impact on ship pose estimation accuracy.

\begin{figure}[t]
\centering
\includegraphics[width=0.9 \linewidth]{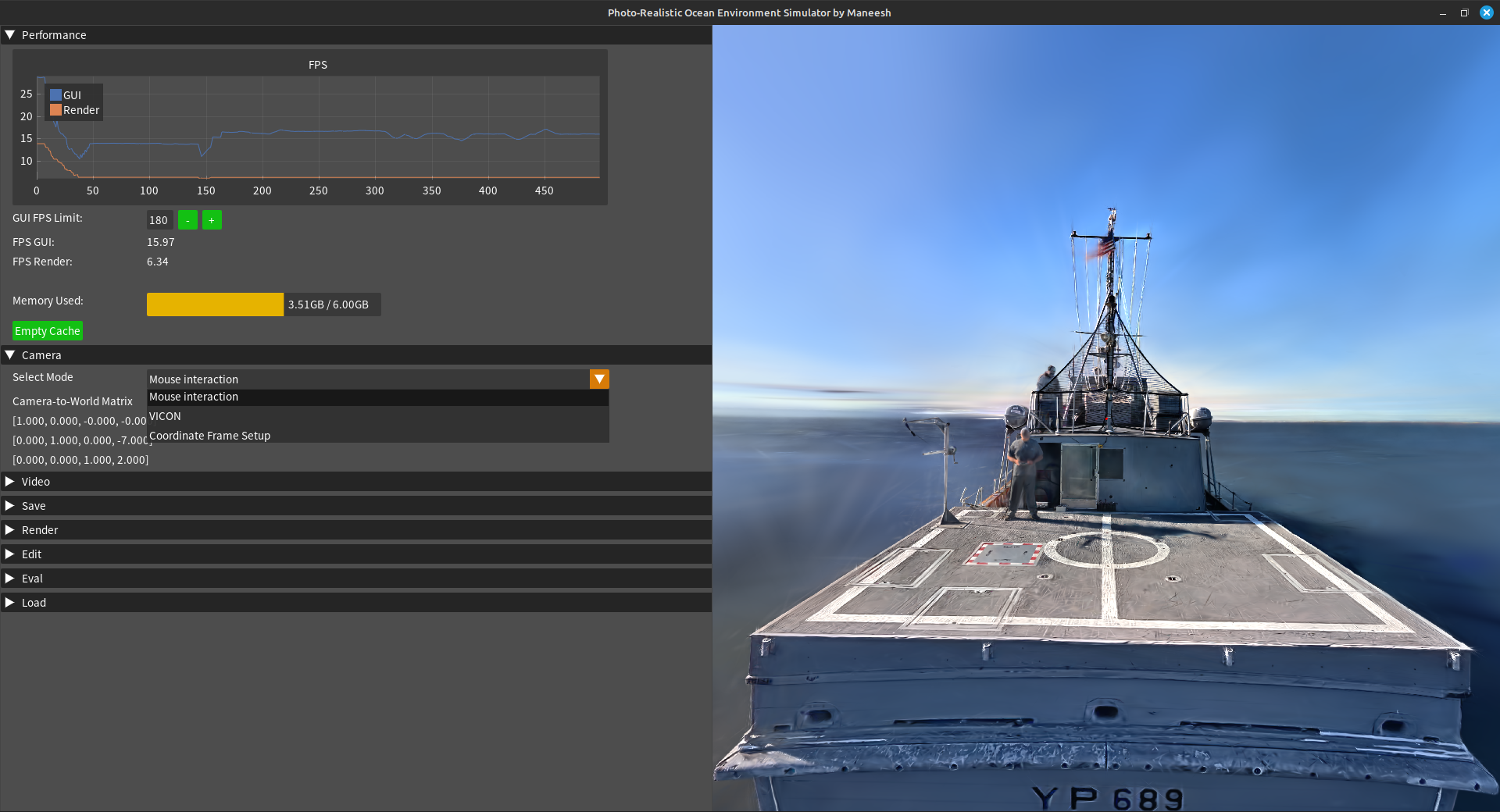}
\caption{The Photo-Realistic Ocean Environment Simulator.}
\label{fig:9}
\end{figure}

To seamlessly integrate the output of 3DGS with the TNN-MO model and other parts of the flight software, we developed custom software named the ``Photo-Realistic Ocean Environment Simulator,'' as shown in \Cref{fig:9}.
In developing the software, we integrated key components from the Splatviz Viewer \cite{barthel2024gaussian}, which is specifically designed for real-time visualization and interaction with 3D Gaussian Splatting (3DGS) scenes.
Built using the Python GUI library PyImGui, the simulator provides an intuitive interface and enhanced functionalities tailored to our application requirements.
It achieves real-time rendering capability, generating images at a resolution of 640 $\times$ 640 with a frame rate ranging from 75 to 110 FPS on a laptop equipped with an NVIDIA GeForce RTX 3060 GPU.

This software has a direct interface with a Vicon motion capture system.
Specifically, the pose of a UAV in the indoor flight experiment area captured by Vicon is transmitted to the 3DGS model such that a photorealistic scene is created, where the Vicon pose in the indoor area corresponds to the pose of the UAV relative to the flight deck in the virtual ocean environment. 
In other words, this system creates a visual scene that makes it appear as if the UAV in the indoor test area is flying in an ocean environment.
The photo-realistic images can be saved up to \qty{60}{FPS} alongside their corresponding Vicon pose data as ground truth.
These data are valuable for testing, validating, and analyzing vision-based models, such as TNN-MO inference, in offline scenarios.

Additionally, the simulator includes an optional real-time 6D pose estimation module that integrates the TNN-MO model to estimate camera poses directly from the rendered images.
This module provides real-time visualization of the keypoints predicted by the TNN-MO model, as shown in \Cref{fig:UAVandTV}.
It is particularly useful for evaluating the tracking performance and robustness of vision-based models.
The software’s real-time 6D pose estimation capability facilitates the development and testing of online UAV state estimators in indoor environments.
By simulating realistic conditions, it increases the likelihood of success while mitigating challenges encountered in real-world scenarios.
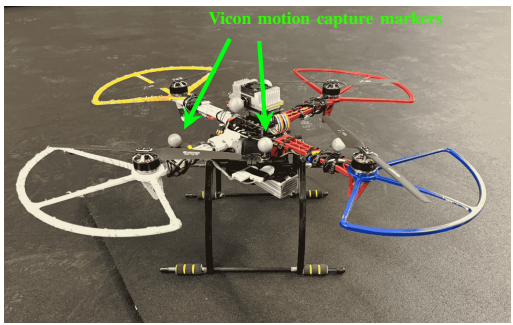
\begin{figure}[b]
	\centering
	\scalebox{0.55}{       
		
		\begin{tikzpicture}[x=0.75pt,y=0.75pt,yscale=-1,xscale=1]
			\draw (339.92,152) node  {\includegraphics[width=348pt,height=217.62pt]{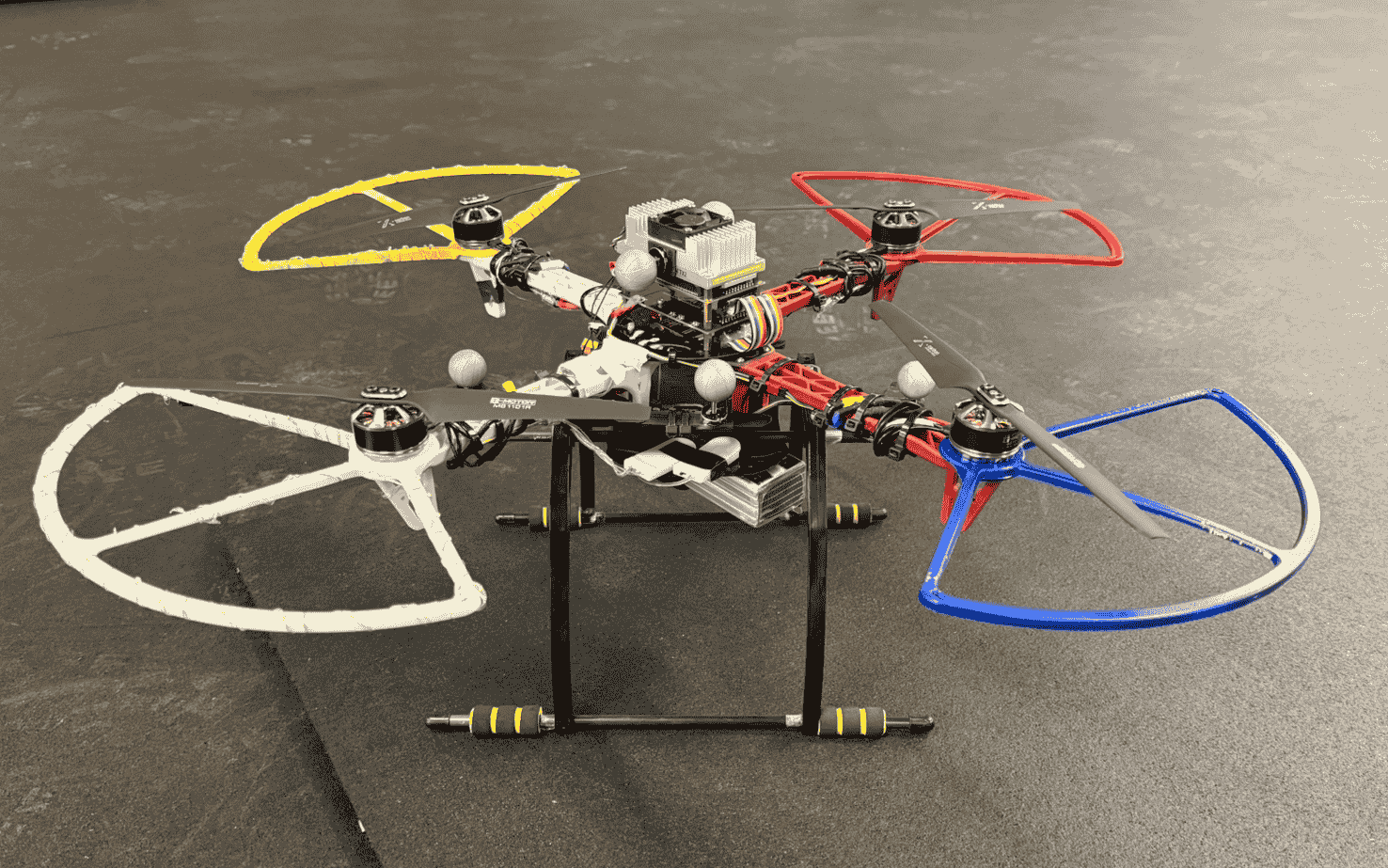}};
			
			\draw [color={rgb, 255:red, 0; green, 255; blue, 18 }  ,draw opacity=1 ][fill={rgb, 255:red, 255; green, 57; blue, 0 }  ,fill opacity=1 ][line width=3]    (314.92,36.92) -- (275.63,114.56) ;
			\draw [shift={(273,120)}, rotate = 296.84] [fill={rgb, 255:red, 0; green, 255; blue, 18 }  ,fill opacity=1 ][line width=0.08]  [draw opacity=0] (18.75,-9.01) -- (0,0) -- (18.75,9.01) -- (12.45,0) -- cycle    ;
			
			\draw [color={rgb, 255:red, 0; green, 255; blue, 18 }  ,draw opacity=1 ][fill={rgb, 255:red, 255; green, 57; blue, 0 }  ,fill opacity=1 ][line width=3]    (341.92,38.92) -- (347.48,115.93) ;
			\draw [shift={(348,122)}, rotate = 265.87] [fill={rgb, 255:red, 0; green, 255; blue, 18 }  ,fill opacity=1 ][line width=0.08]  [draw opacity=0] (18.75,-9.01) -- (0,0) -- (18.75,9.01) -- (12.45,0) -- cycle    ;

			\draw (294,11) node [anchor=north west][inner sep=0.75pt]   [align=left] {\textbf{{\large \textcolor[rgb]{0.18,1,0}{Vicon motion capture markers}}}};
			
		\end{tikzpicture}
	}
	\caption{Quadrotor with Vicon motion capture markers used for indoor flight experiments~\cite{yu2024modular}.}
	\label{fig:10}
\end{figure}

\section{Vision-In-The-Loop Simulation}\label{sec:Sim}

\subsection{Flight Hardware and Software}

The UAV, shown in \Cref{fig:10}, consists of a VectorNav VN100 IMU, T-Motor 700KV brushless motors, and MS1101 propellers, controlled by MikroKopter BL-Ctrl v2 ESCs to ensure stable and precise flight.
A custom PCB handles power distribution, communication interfaces (I2C, UART), and sensor integration, powered by a 14.8V 6000mAh Li-Po battery.

The flight software runs on an onboard NVIDIA Jetson TX2, employing a multi-threaded architecture for real-time sensor processing, state estimation, and motor control.
A delayed Kalman filter integrates IMU and GPS data for precise state estimation~\cite{KaniDelKal}, while an adaptive geometric controller ensures robust trajectory tracking~\cite{GamLeeAJDSMC22,LeeLeoPICDC10}.
A GTKmm-based GUI enables real-time monitoring, mission planning, and data visualization, with Wi-Fi ensuring seamless communication between the UAV and the ground station.

For indoor flight experiments, position and attitude measurements are obtained via a Vicon motion capture system with twelve Valkyrie VK-8 cameras, which can achieve sub-millimeter accuracy (often around 0.1 mm to 1 mm) under optimal conditions, transmitting data at 200 Hz over Wi-Fi.
For outdoor operations, a SwiftNav Piksi Multi RTK GPS provides centimeter-level positioning.

\subsection{Vision-In-The-Loop Test Framework}

\begin{figure}[b]
\centering
\scalebox{0.48}{
\begin{tikzpicture}[x=0.75pt,y=0.75pt,yscale=-1,xscale=1]
	\draw  [fill={rgb, 255:red, 65; green, 117; blue, 5 }  ,fill opacity=0.27 ][line width=1.5]  (340,40) -- (630,40) -- (630,110) -- (490,110) -- (490,240) -- (340,240) -- cycle ;

	\draw  [fill={rgb, 255:red, 155; green, 155; blue, 155 }  ,fill opacity=0.3 ][line width=1.5]  (30,130) -- (200,130) -- (200,270) -- (30,270) -- cycle ;

	\draw  [fill={rgb, 255:red, 74; green, 144; blue, 226 }  ,fill opacity=0.51 ] (40,170) -- (190,170) -- (190,260) -- (40,260) -- cycle ;

	\draw  [fill={rgb, 255:red, 126; green, 211; blue, 33 }  ,fill opacity=0.3 ] (530,50) -- (620,50) -- (620,90) -- (530,90) -- cycle ;

	\draw  [fill={rgb, 255:red, 126; green, 211; blue, 33 }  ,fill opacity=0.3 ] (370,50) -- (460,50) -- (460,90) -- (370,90) -- cycle ;

	\draw  [fill={rgb, 255:red, 126; green, 211; blue, 33 }  ,fill opacity=0.3 ] (380,190) -- (460,190) -- (460,230) -- (380,230) -- cycle ;

	\draw  [fill={rgb, 255:red, 248; green, 189; blue, 28 }  ,fill opacity=0.42 ] (550,190) -- (630,190) -- (630,230) -- (550,230) -- cycle ;

	\draw    (660,70) -- (620,70) ;

	\draw    (460,70) -- (528,70) ;
	\draw [shift={(530,70)}, rotate = 180] [fill={rgb, 255:red, 0; green, 0; blue, 0 }  ][line width=0.08]  [draw opacity=0] (12,-3) -- (0,0) -- (12,3) -- cycle    ;

	\draw    (650,70) -- (650,210) -- (632,210) ;
	\draw [shift={(630,210)}, rotate = 360] [fill={rgb, 255:red, 0; green, 0; blue, 0 }  ][line width=0.08]  [draw opacity=0] (12,-3) -- (0,0) -- (12,3) -- cycle    ;

	\draw    (550,210) -- (462,210) ;
	\draw [shift={(460,210)}, rotate = 360] [fill={rgb, 255:red, 0; green, 0; blue, 0 }  ][line width=0.08]  [draw opacity=0] (12,-3) -- (0,0) -- (12,3) -- cycle    ;

	\draw    (430,190) -- (430,92) ;
	\draw [shift={(430,90)}, rotate = 90] [fill={rgb, 255:red, 0; green, 0; blue, 0 }  ][line width=0.08]  [draw opacity=0] (12,-3) -- (0,0) -- (12,3) -- cycle    ;

	\draw  [fill={rgb, 255:red, 189; green, 16; blue, 224 }  ,fill opacity=0.48 ][dash pattern={on 4.5pt off 4.5pt}] (220,190) -- (300,190) -- (300,230) -- (220,230) -- cycle ;

	\draw  [dash pattern={on 4.5pt off 4.5pt}]  (300,210) -- (371,210) -- (378,210) ;
	\draw [shift={(380,210)}, rotate = 180] [fill={rgb, 255:red, 0; green, 0; blue, 0 }  ][line width=0.08]  [draw opacity=0] (12,-3) -- (0,0) -- (12,3) -- cycle    ;

	\draw  [fill={rgb, 255:red, 128; green, 218; blue, 255 }  ,fill opacity=0.52 ][line width=1.5]  (40,20) -- (180,20) -- (180,100) -- (40,100) -- cycle ;

	\draw    (130,210) -- (218,210) ;
	\draw [shift={(220,210)}, rotate = 180] [fill={rgb, 255:red, 0; green, 0; blue, 0 }  ][line width=0.08]  [draw opacity=0] (12,-3) -- (0,0) -- (12,3) -- cycle    ;

	\draw  [dash pattern={on 4.5pt off 4.5pt}]  (140,60) -- (260,60) -- (260,188) ;
	\draw [shift={(260,190)}, rotate = 270] [fill={rgb, 255:red, 0; green, 0; blue, 0 }  ][line width=0.08]  [draw opacity=0] (12,-3) -- (0,0) -- (12,3) -- cycle    ;

	\draw    (590,230) -- (590,250) -- (90,250) -- (90,232) ;
	\draw [shift={(90,230)}, rotate = 90] [fill={rgb, 255:red, 0; green, 0; blue, 0 }  ][line width=0.08]  [draw opacity=0] (12,-3) -- (0,0) -- (12,3) -- cycle    ;

	\draw  [fill={rgb, 255:red, 144; green, 19; blue, 254 }  ,fill opacity=0.25 ] (70,40) -- (140,40) -- (140,80) -- (70,80) -- cycle ;

	\draw  [fill={rgb, 255:red, 80; green, 227; blue, 194 }  ,fill opacity=0.46 ] (50,190) -- (130,190) -- (130,230) -- (50,230) -- cycle ;

	\draw  [dash pattern={on 4.5pt off 4.5pt}]  (330,270) -- (368.23,270) -- (380,270) ;

	\draw  [color={rgb, 255:red, 65; green, 117; blue, 5 }  ,draw opacity=1 ][line width=2.25]  (310,120) -- (310,10) -- (670,10) -- (670,300) -- (20,300) -- (20,120) -- cycle ;

	\draw (535,62) node [anchor=north west][inner sep=0.75pt]  [font=\large] [align=left] {Quadrotor};

	\draw (384,63) node [anchor=north west][inner sep=0.75pt]  [font=\large] [align=left] {Control};

	\draw (383.4,203) node [anchor=north west][inner sep=0.75pt]  [font=\large] [align=left] {Estimator};

	\draw (560,204) node [anchor=north west][inner sep=0.75pt]  [font=\large] [align=left] {Vicon};

	\draw (347,132.4) node [anchor=north west][inner sep=0.75pt]  [font=\large]  {$x,\ R,\ v,\ \Omega $};

	\draw (468,52) node [anchor=north west][inner sep=0.75pt]  [font=\normalsize] [align=left] {\textit{Throttle }};

	\draw (501,195.4) node [anchor=north west][inner sep=0.75pt]  [font=\large]  {$x,\ R$};

	\draw (224,203) node [anchor=north west][inner sep=0.75pt]  [font=\large] [align=left] {TNN-MO};

	\draw (301,190.4) node [anchor=north west][inner sep=0.75pt]  [font=\large]  {$\hat{x} ,\ \hat{R}$};

	\draw (66,204) node [anchor=north west][inner sep=0.75pt]  [font=\large] [align=left] {3DGS};

	\draw (131,178) node [anchor=north west][inner sep=0.75pt]  [font=\normalsize] [align=left] {\begin{minipage}[lt]{41.82pt}\setlength\topsep{0pt}
			\begin{center}
				\textit{Renderd}\\\textit{Image}
			\end{center}
			
	\end{minipage}};
	\draw (42,135) node [anchor=north west][inner sep=0.75pt]  [font=\normalsize] [align=left] {\begin{minipage}[lt]{101.32pt}\setlength\topsep{0pt}
			\begin{center}
				Photo realistic Scene \\Generator
			\end{center}
			
	\end{minipage}};
	\draw (41,274) node [anchor=north west][inner sep=0.75pt]   [align=left] {\textbf{Virtual Environment}};
	\draw (44,2) node [anchor=north west][inner sep=0.75pt]   [align=left] {\textbf{Real Environment}};
	\draw (191,47) node [anchor=north west][inner sep=0.75pt]  [font=\normalsize] [align=left] {\begin{minipage}[lt]{54.85pt}\setlength\topsep{0pt}
			\begin{center}
				\textit{Real Image}
			\end{center}
			
	\end{minipage}};
	\draw (80,52) node [anchor=north west][inner sep=0.75pt]  [font=\large] [align=left] {Camera};
	\draw (391,264) node [anchor=north west][inner sep=0.75pt]  [font=\small] [align=left] {\begin{minipage}[lt]{150.46pt}\setlength\topsep{0pt}
			\begin{center}
				\textit{To be implemented in Nvidia Jetson }
			\end{center}
			
	\end{minipage}};
	\draw (391,21) node [anchor=north west][inner sep=0.75pt]   [align=left] {\textbf{UAV / Nvidia Jetson }};
	\draw (165,305) node [anchor=north west][inner sep=0.75pt]  [font=\Large,color={rgb, 255:red, 0; green, 0; blue, 0 }  ,opacity=1 ] [align=left] {Vision-In-The-Loop Simulation Framework};

\end{tikzpicture}

}
\caption{Overview of the Vision-In-Loop framework for state estimation and control of the UAV.}
\label{fig:11}
\end{figure}
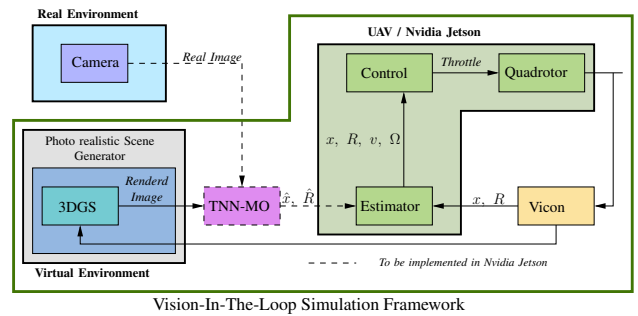

The vision-in-the-loop simulation framework integrates flight hardware, software, and the TNN-MO model for monocular pose estimation, enabling comprehensive testing of autonomous UAV flight algorithms with integrated vision-based pose estimation schemes.
An overview of the proposed simulation framework is illustrated in \Cref{fig:11}.

For a quadrotor UAV flying in an indoor environment, its pose is measured by an external Vicon motion capture system, providing the position and attitude pair $(x,R)\in\Re^3\times\SO$, where the special orthogonal group is denoted by $\SO={R\in\Re^{3\times 3},|, R^T R = I_{3\times 3}, \mathrm{det}[R]=1}$.
The Vicon measurements are considered ground truth.

This data is then transferred to the estimation thread of the onboard flight software, where it is integrated with IMU measurements to estimate the UAV’s complete state, namely $(x,R, v, \Omega)\in\Re^3\times\SO\times \Re^3\times\Re^3$, which includes the additional linear velocity $v$ and angular velocity $\Omega$.
Based on the estimated state, the control thread of the flight software computes throttle commands for each rotor, ensuring the UAV follows the desired trajectory safely.
Both the estimator and control algorithms are optimized for deployment on an onboard Nvidia Jetson platform, ensuring real-time applicability.
\begin{figure}[t]
	\centering
	\includegraphics[width=1 \linewidth]{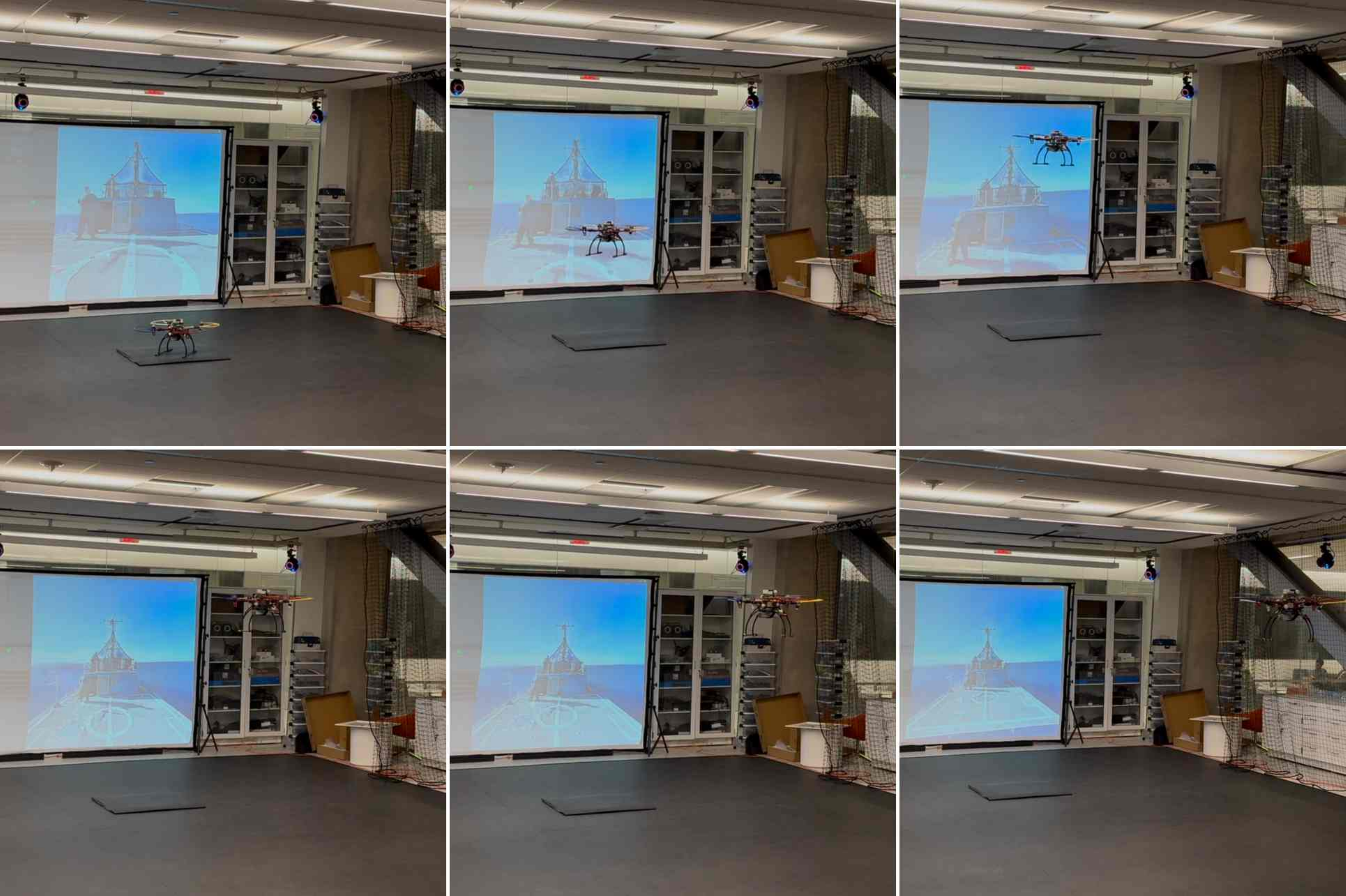}
	\caption{Vision-in-the-loop flight experiments: demonstrating real-time photo-realistic image generation from software displayed on a TV with various UAV poses.
	A video of our experiments can be found a
	\href{https://youtu.be/xYWxhp70PdA}{https://youtu.be/xYWxhp70PdA}.}
	\label{fig:UAVandTV}
\end{figure}
Meanwhile, the pose measured by the Vicon motion capture system is sent to the proposed 3DGS framework to generate photo-realistic images corresponding to $(x,R)$ in the virtual environment.
These images are then fed into the TNN-MO model to estimate the UAV’s 6D pose relative to the ship $(\hat{x}, \hat{R})$, as detailed in \Cref{sec:PRVE}.
\begin{figure}[b]
	\centering
	\scalebox{0.46}{
		\tikzset{every picture/.style={line width=0.75pt}}      
		
		\begin{tikzpicture}[x=0.75pt,y=0.75pt,yscale=-1,xscale=1] 
			\draw (345.96,264.72) node  {\includegraphics[width=520.44pt,height=391.09pt]{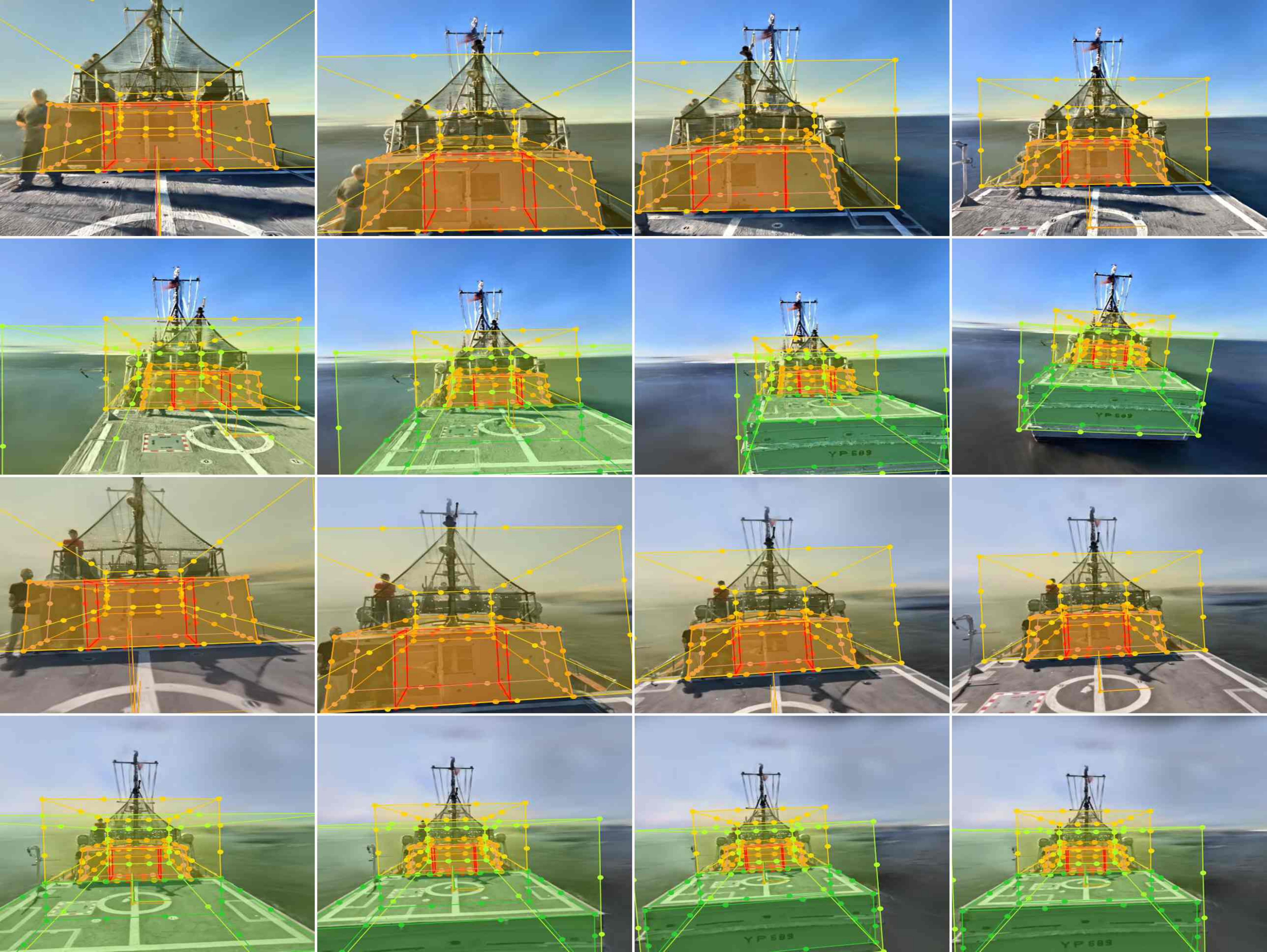}};
			
			\draw (6,7) node [anchor=north west][inner sep=0.75pt]   [align=left] {\textcolor[rgb]{1,1,1}{{\LARGE \textbf{Test 1 (in Environment 1)}}}};
			
			\draw (6,267) node [anchor=north west][inner sep=0.75pt]   [align=left] {\textcolor[rgb]{1,1,1}{{\LARGE \textbf{Test 2 (in Environment 2)}}}};
		\end{tikzpicture}
		
	}
	\caption{Top two rows present 3DGS environment 1 (Test 1), and the bottom two rows show 3DGS environment 2 (Test 2). The visualization highlights keypoints predicted by the TNN-MO model from rendered images, alongside real-time photo-realistic image generation and 6D pose estimation during indoor flights.}
	\label{fig:13}
\end{figure}
The TNN-MO model operates alongside the Photorealistic Ocean Environment Simulator at a frequency of 10 Hz in a ground station, with estimated poses recorded for analysis.

In the current setup, the UAV is controlled based on the pose measured by the Vicon system.
For future implementations, the Vicon system will be replaced with TNN-MO-based pose estimations $(\hat{x}, \hat{R})$, fused with IMU measurements to estimate the full state $(x, R, v, \Omega)$ for the flight controller, as indicated by the dashed line in \Cref{fig:11}.

\subsection{Flight Test Results}
\begin{table*}[t]
	\centering
	\caption{Accuracy of 6D Pose Estimation for TNN-MO Model Under different photo-realistic environments}
	\label{tab:Exp}
	\begin{tabular}{ccccc}
		\toprule
		\toprule
		Image Type  & Max Range, $L$ (m)  & MAE / std. $\sigma$ of Pos. (m) & MAE/$L$ (\%) & MAE of Rot. (deg)\\
		\midrule
		Test 1 in Environment 1   & 11.9 & 0.105 / 0.304 & 0.88 & 1.78\\
		Test 2 in Environment 2 & 10.1  & 0.089 / 0.217 & 0.76 & 1.54\\
		\bottomrule
	\end{tabular}
\end{table*} 

To evaluate the performance and effectiveness of the vision-in-the-loop simulation, a series of flight experiments were conducted in an indoor laboratory environment, as shown in \Cref{fig:UAVandTV}.
The UAV was tasked with executing takeoff, path-following, and landing maneuvers.
The pose estimates produced by the TNN-MO model were analyzed and compared with the ground truth Vicon pose data to assess accuracy and robustness.
\begin{figure}[b!] 
	\centering
	\includegraphics[width=1\linewidth]{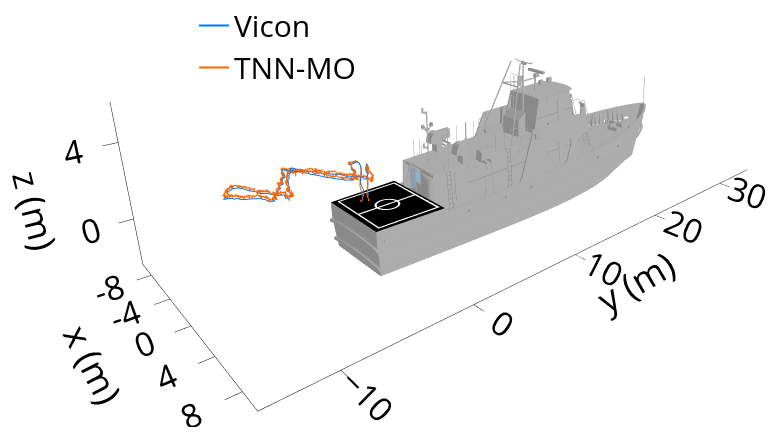}
	\caption{3D view of the ship for Test 1}
	\label{fig:14}
\end{figure}
\begin{figure}[b!] 
	\centering
	\includegraphics[width=0.95\linewidth]{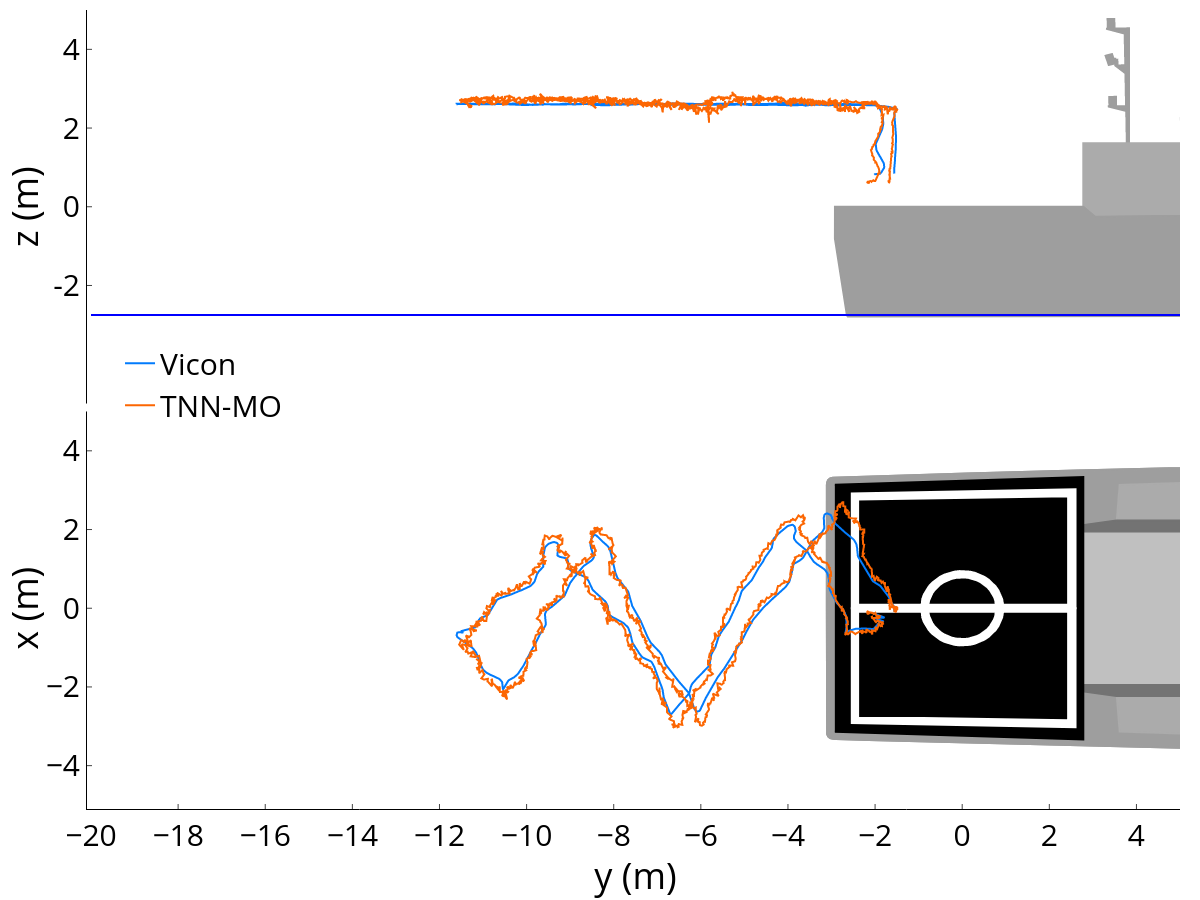}
	\caption{Top and side view of the ship for Test 1}
	\label{fig:15}
\end{figure}

\begin{figure}[t] 
	\begin{subfigure}
		\centering
		\includegraphics[width=1\linewidth]{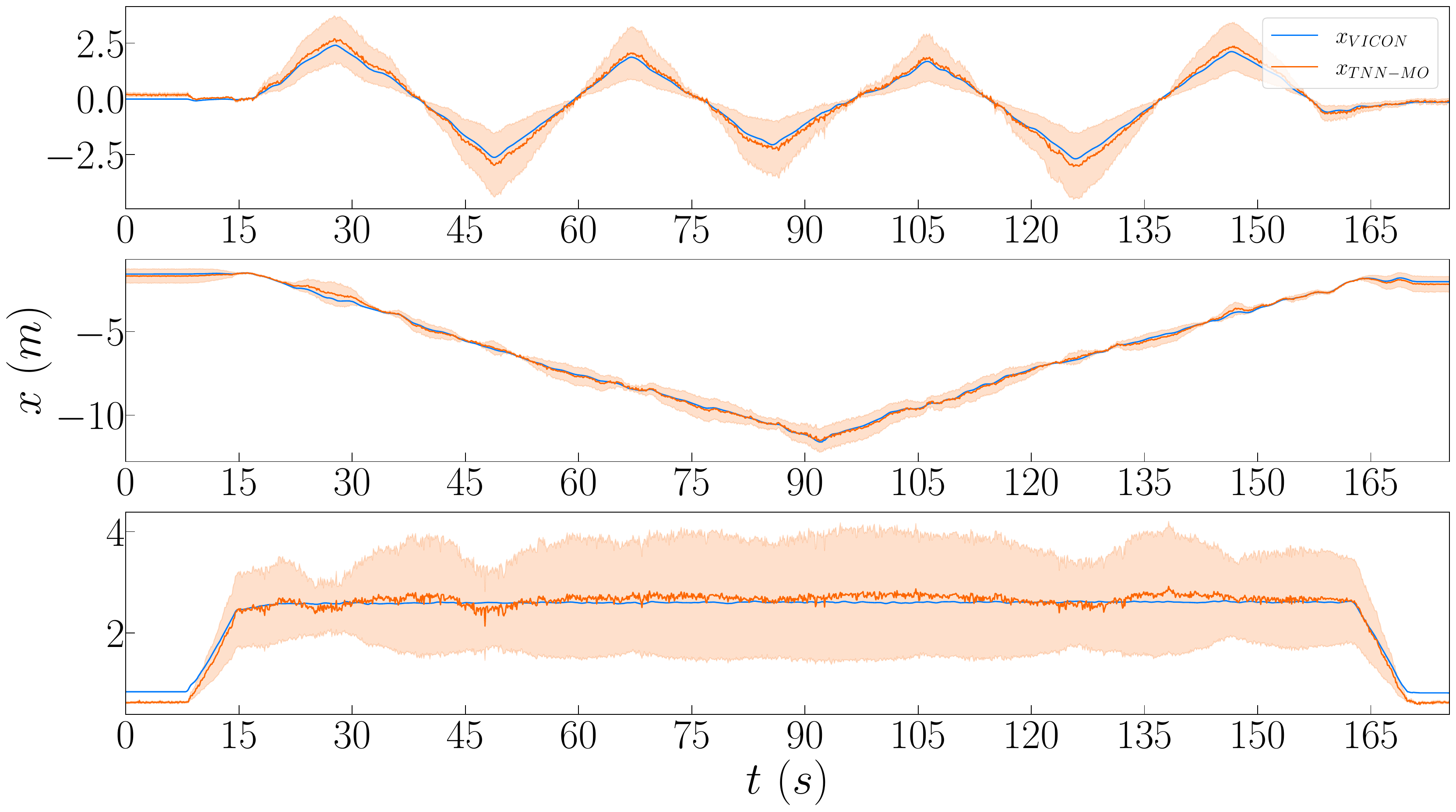}
		\caption{Position, $x$ for Test 1: the estimated position with uncertainty (orange) is compared against the Vicon position measurements (blue).}
		\label{fig:16}
	\end{subfigure}
	\vspace{4mm}
	\begin{subfigure}
		\centering
		\includegraphics[width=1\linewidth]{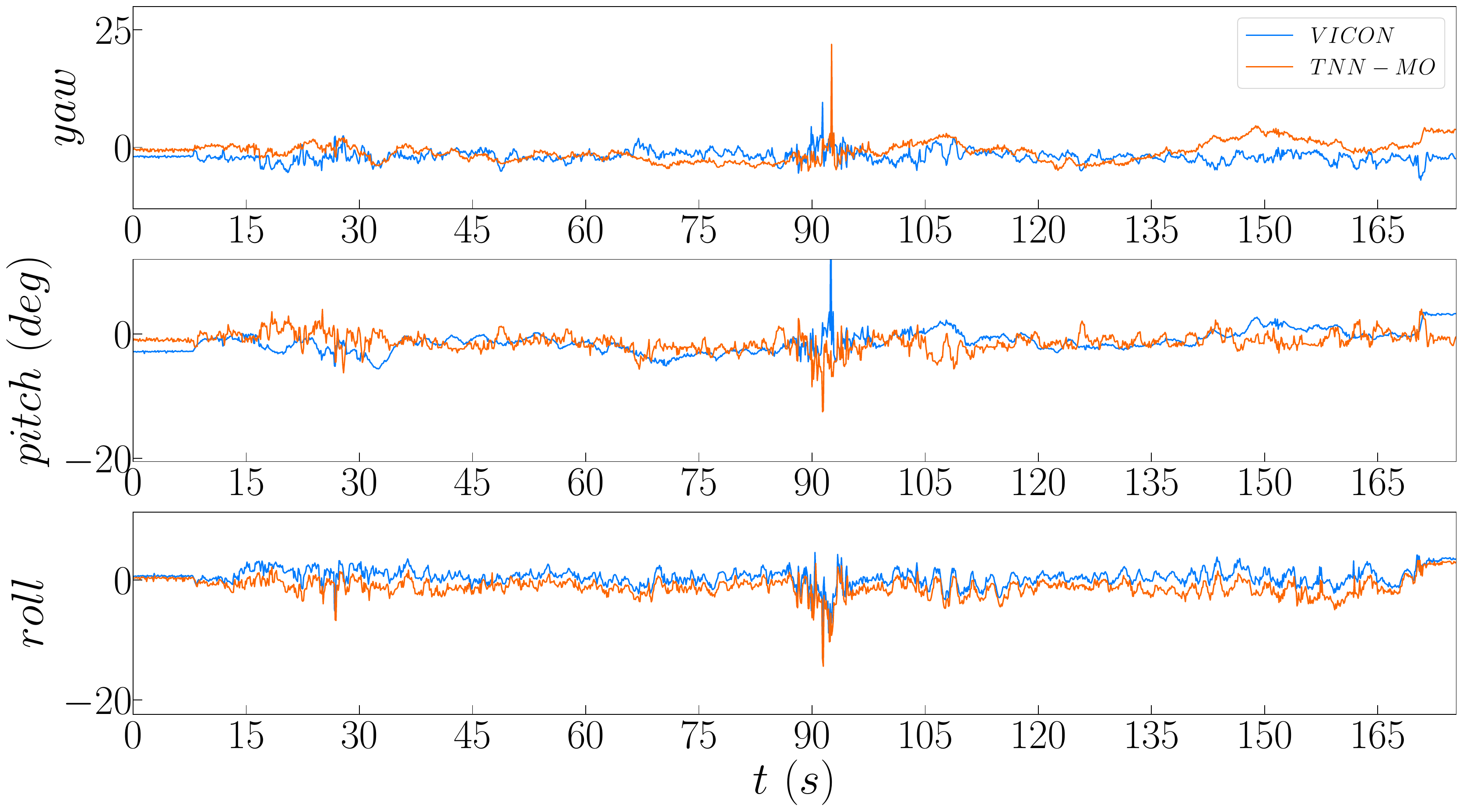}
		\caption{Attitude, $ypr$ for Test 1: the estimated attitude (orange) is compared against the Vicon attitude (blue).}
		\label{fig:18}
	\end{subfigure}
\end{figure}

We consider two test scenarios with different virtual environments, as shown in \Cref{fig:13}.
The first scenario (Test 1) involves a zig-zag flight path on a bright day.
The resulting actual flight trajectory (blue) and the estimated trajectory (orange) are illustrated with respect to the ship in \Cref{fig:14,fig:15}.
The position $x$ is plotted with respect to time in \Cref{fig:16} along with a $2\sigma$ bound, and the attitude error is presented in terms of the errors in the Euler angles in \Cref{fig:18}.
\begin{figure}[!b]
	\centering
	\includegraphics[width=1\linewidth]{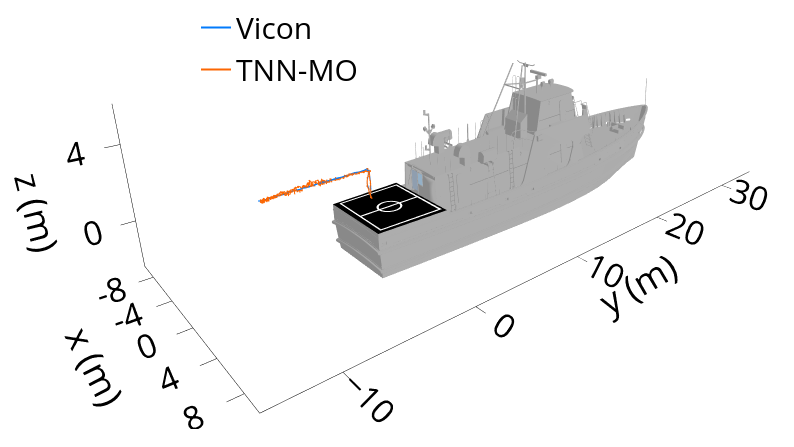}
	\caption{3D view of the ship for Test 2}
	\label{fig:19}
\end{figure}
These results show that the estimated trajectories are consistent with the true pose measured by Vicon.
Specifically, as summarized in \Cref{tab:Exp}, the model achieves a mean absolute error (MAE) of \SI{0.105}{\meter} for position, with a standard deviation (std.) of \SI{0.3}{\meter}, and the position error is about to 0.88\% of the maximum operational range (\SI{12}{\meter}).
\begin{figure}[!t]
	\centering
	\includegraphics[width=0.95\linewidth]{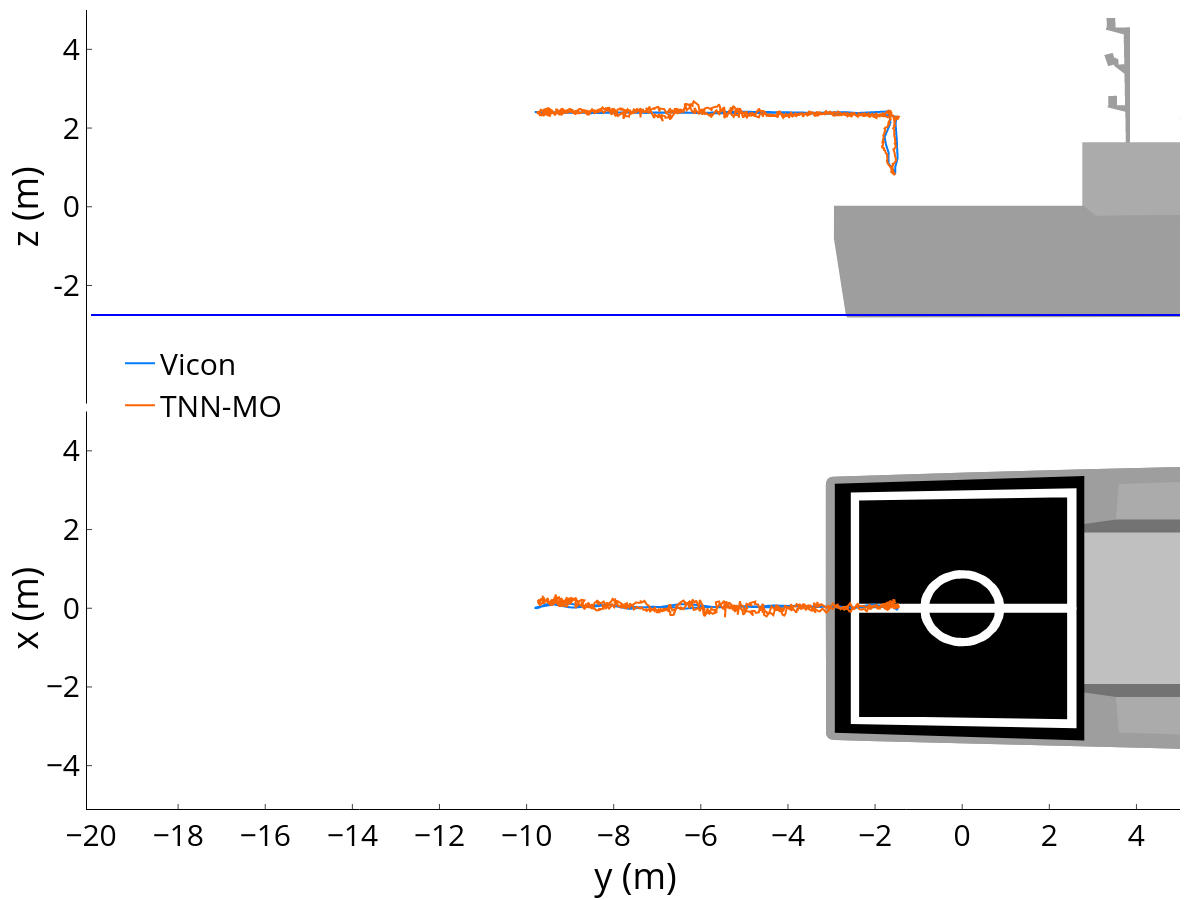}
	\caption{Top and side view of the ship for Test 2}
	\label{fig:20}
\end{figure}

\begin{figure}[!t]
	\centering
	\includegraphics[width=1\linewidth]{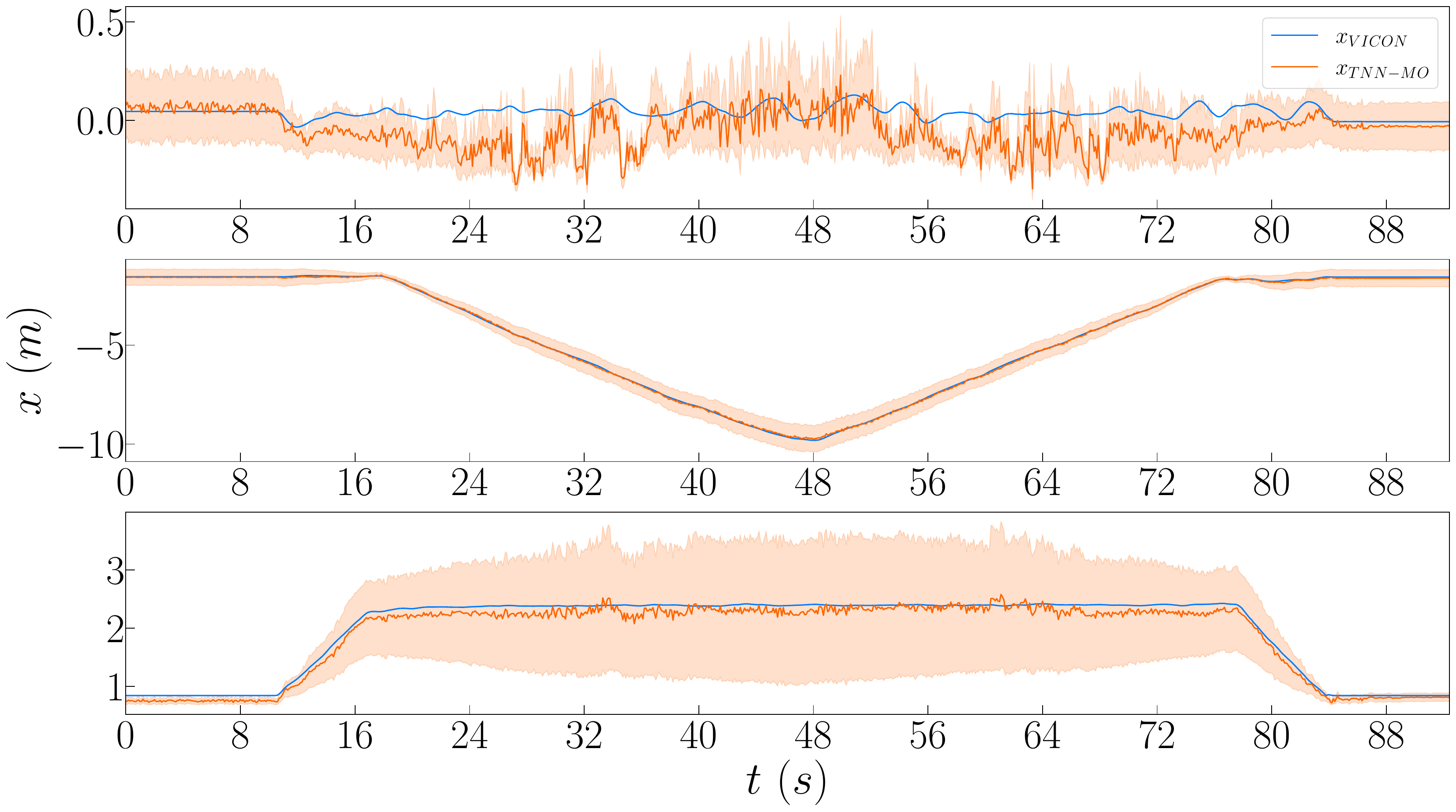}
	\caption{Position, $x$ for Test 2: the estimated position with uncertainty (orange) is compared against the Vicon position measurements (blue).}
	\label{fig:21}
\end{figure}

\begin{figure}[!t]
	\centering
	\includegraphics[width=1\linewidth]{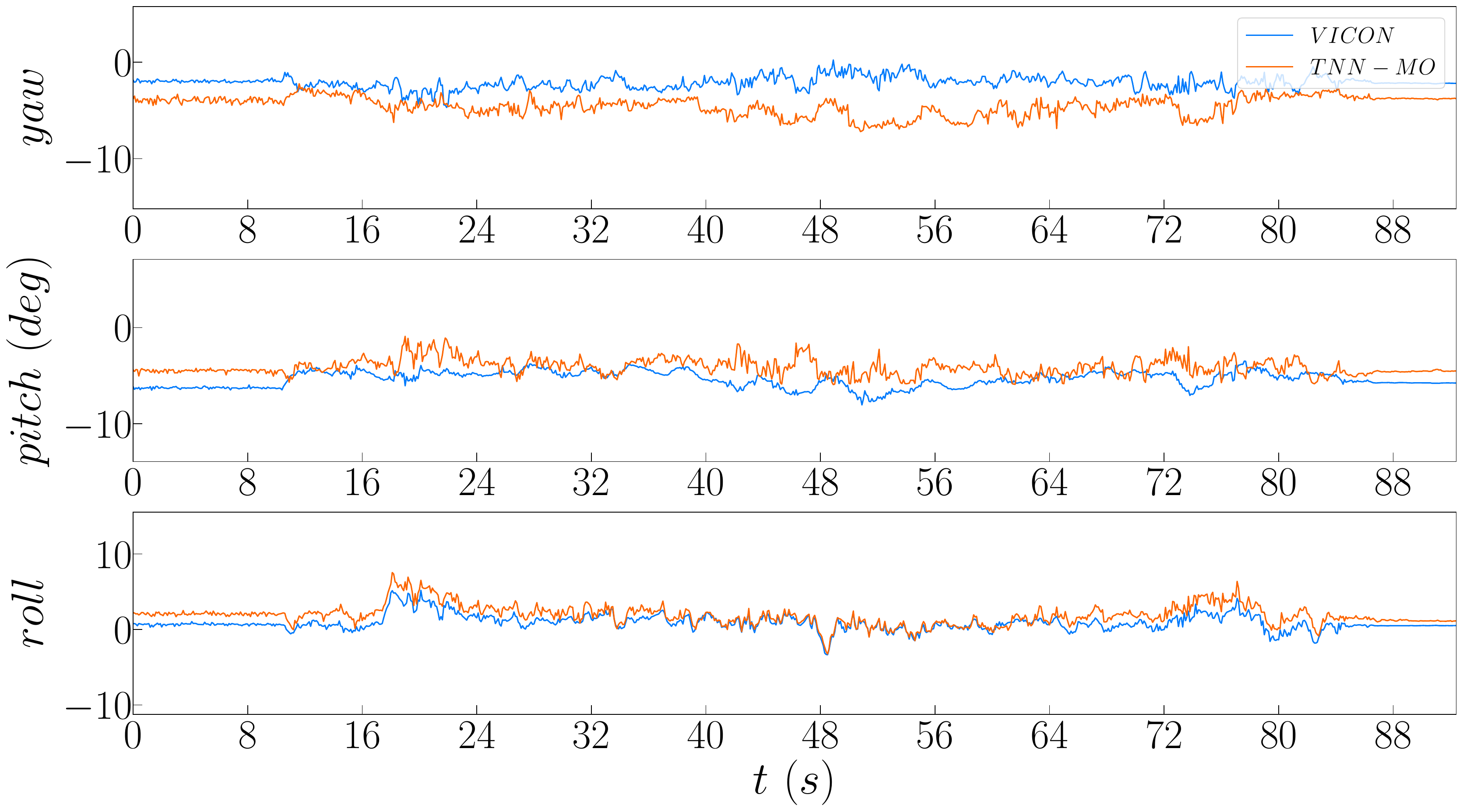}
	\caption{Attitude, $ypr$ for Test 2: the estimated attitude (orange) is compared against the Vicon attitude (blue).}
	\label{fig:23}
\end{figure}

The MAE of the attitude is \qty{1.78}{\degree}.
These values are consistent with other state-of-the-art techniques in vision-based localization.
More importantly, these errors result from both the generation of the photo-realistic scene with 3DGS and the estimation of the pose with TNN-MO.

The second scenario (Test 2) involves a straight path on a cloudy day.
The corresponding results, illustrated in \Cref{fig:19,fig:20,fig:21,fig:23}, and the errors summarized in \Cref{tab:Exp}, are consistent with those of Test 1 or slightly improved.
These results underscore the model’s consistency in delivering accurate pose estimations under varying simulated maritime conditions.

%
These findings demonstrate the feasibility of validating vision-based control and estimation algorithms in an indoor environment under realistic maritime conditions, eliminating the need for costly and high-risk real-world experiments.

Once robust vision-based state estimation algorithms are successfully validated within the vision-in-the-loop simulation framework, the system can transition to real-world experiments.
By leveraging this framework, researchers gain a cost-effective and scalable platform for rigorous validation of deep monocular pose estimation models.
This approach accelerates the development of reliable UAV systems tailored for maritime applications while systematically addressing challenges such as environmental variability and sensor noise.
Consequently, the framework significantly advances the deployment of vision-driven autonomous systems in complex, dynamic environments.

\section{Conclusions}

This study presents a vision-in-the-loop simulation framework for validating deep monocular pose estimation and vision-based control of UAVs in maritime environments.
By leveraging 3D Gaussian Splatting (3DGS), a photo-realistic virtual ocean environment is constructed, enabling indoor testing of UAV flight algorithms without the need for costly real-world deployments.
The proposed Transformer Neural Network Multi-Object (TNN-MO) model achieves mean absolute position errors of \SI{0.105}{\meter} (Test 1) and \SI{0.089}{\meter} (Test 2), with attitude errors of \SI{1.8}{\degree} and \SI{1.54}{\degree}, respectively, across two distinct 3DGS environments.
These results demonstrate the framework’s ability to simulate dynamic maritime conditions while ensuring accurate 6D pose estimation.

Future work will focus on integrating TNN-MO with an IMU through the onboard estimation scheme to validate the complete flight hardware and software without relying on an external motion capture system,
as well as conducting real-world flight experiments.

\section*{Acknowledgment}
This research was conducted on the US Naval Academy’s research vessel YP689, and we extend our gratitude to the US Naval Academy and the crew of YP689. 

\bibliography{ICUAS25}
\bibliographystyle{IEEEtran}

\end{document}